%% file: proposal.tex
\documentclass[letterpaper, 10 pt, conference]{ieeeconf}  

\IEEEoverridecommandlockouts                              

\usepackage{graphics} 
\usepackage{epsfig} 
\usepackage{mathptmx} 
\usepackage{times} 
\usepackage{amsmath}
\usepackage{bm}
\usepackage{amssymb}  
\usepackage{booktabs}
\usepackage{gensymb}
\usepackage{multirow}
\usepackage{subcaption}
\usepackage{stfloats}
\usepackage{array}
\usepackage{siunitx} 
\usepackage{graphicx} 
\usepackage[ruled,vlined]{algorithm2e}
\usepackage[section]{placeins}
\usepackage{utfsym}
\DeclareMathAlphabet\mathbfcal{OMS}{cmsy}{b}{n}

\title{\LARGE \bf
Metric, inertially aligned monocular state estimation via kinetodynamic priors
}

\author{ Jiaxin Liu$^{1*}$, Min Li$^{1*}$, Wanting Xu$^{1}$, Liang Li$^{2}$, Jiaqi Yang$^{1}$ and Laurent Kneip$^{1}$
\thanks{$^{*}$Two authors contribute equally in this work}
\thanks{$^{1}$Jiaxin Liu, Min Li, Wanting Xu, Jiaqi Yang and Laurent Kneip are with School of Information Science and Technology, ShanghaiTech University, Shanghai, China. \{liujx2024, limin1, xuwt, yangjq, lkneip\}@shanghaitech.edu.cn}%
\thanks{$^{2}$Liang Li is with the Department of Mechanical Engineering, The University of Hongkong, Hongkong S.A.R., China. llihku@connect.hku.hk} %
\thanks {Corresponding author: Laurent Kneip, lkneip@shanghaitech.edu.cn}
}

\begin{document}

\maketitle
\thispagestyle{empty}
\pagestyle{empty}

\input{sec/main/0_Abstract}    
\input{sec/main/1_Introduction}
\input{sec/main/2_Relatedwork}

\input{sec/main/3_Methodology}

\input{sec/main/4_Experiments}

\input{sec/main/6_Ablation}

\input{sec/main/5_Conclusion}

\input{sec/main/7_Acknowledgement}
\input{sec/appendix/appendix_icra}

{
   \small
   \bibliographystyle{IEEEtran}
   \bibliography{IEEEexample}
}

\end{document}

%% file: sec/main/0_Abstract.tex
\begin{abstract}

Accurate state estimation for flexible robotic systems poses significant challenges, particularly for platforms with dynamically deforming structures that invalidate rigid-body assumptions. This paper addresses this problem and enables the extension of existing rigid-body pose estimation methods to non-rigid systems. Our approach integrates two core components: first, we capture elastic properties using a deformation-force model, efficiently learned via a Multi-Layer Perceptron; second, we resolve the platform's inherently smooth motion using continuous-time B-spline kinematic models. By continuously applying Newton's Second Law, our method formulates the relationship between visually-derived trajectory acceleration and predicted deformation-induced acceleration. We demonstrate that our approach not only enables robust and accurate pose estimation on non-rigid platforms, but also shows that the properly modeled platform physics allow for the recovery of inertial sensing properties. We validate this feasibility on a simple spring-camera system, showing how it robustly resolves the typically ill-posed problem of metric scale and gravity recovery in monocular visual odometry.

\end{abstract}

%% file: sec/main/1_Introduction.tex
\section{Introduction}
\label{sec:introduction}

\begin{figure}[t]

    \includegraphics[width=0.5\textwidth]{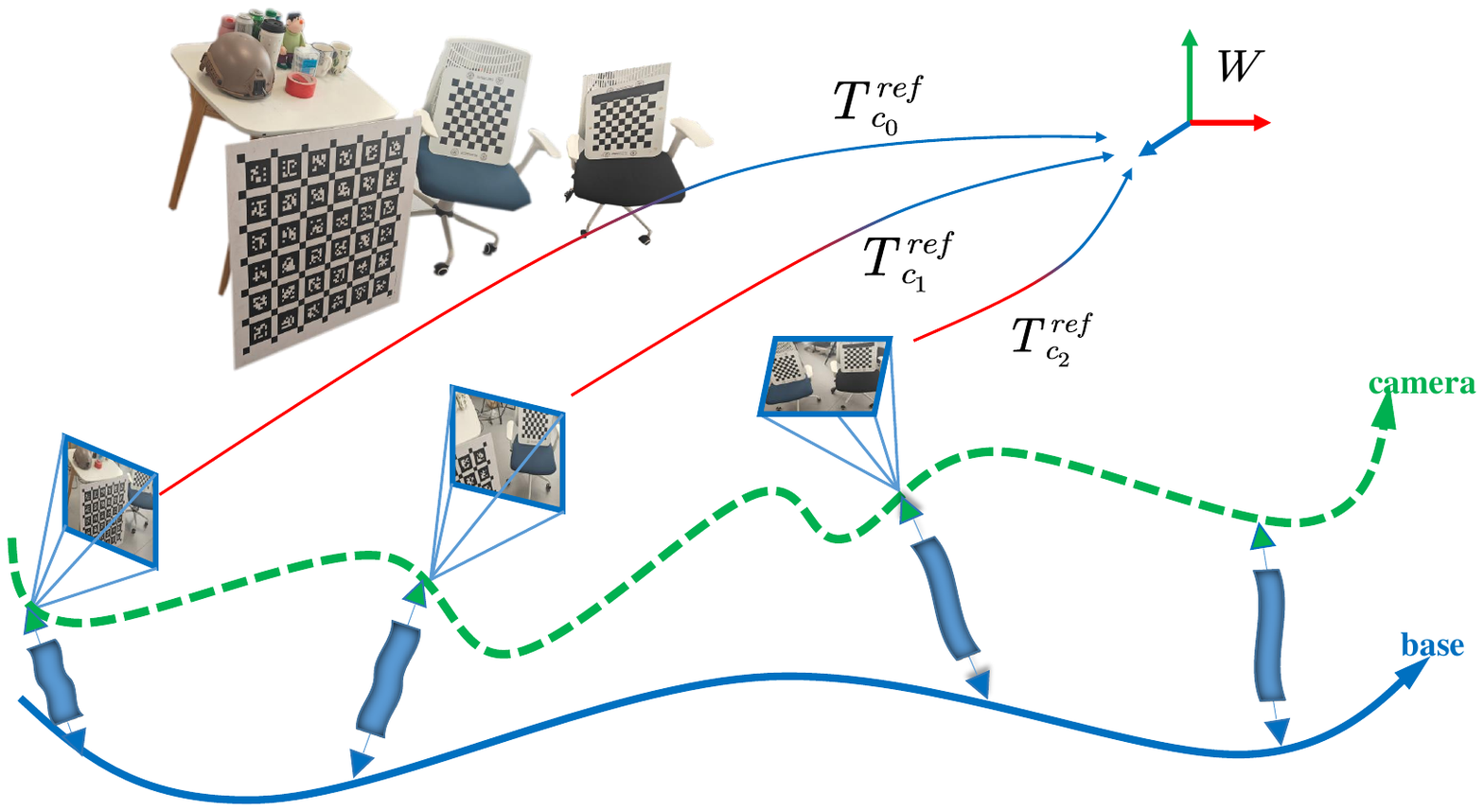}
\vspace{-2 cm}
    \caption{Non-rigid monocular camera system demonstrating trajectory divergence. Unlike rigid systems, the deformable connection induces camera oscillations (green) distinct from the platform's motion (blue).}
\vspace{-0.6 cm}
    \label{fig:nonrigid_system}

\end{figure}
Accurate environmental perception and precise self-pose estimation are paramount for autonomous navigation, human-robot collaboration, and complex robotic tasks. Traditionally, these capabilities have relied on rigid-body assumptions, simplifying multi-sensor fusion and motion estimation. However, the burgeoning field of soft robotics and flexible systems—such as compliant manipulators and UAVs—challenges this paradigm \cite{chen2025survey}. While valued for their inherent safety, enhanced dexterity, and potential for lower manufacturing costs, these non-rigid systems leverage structural deformation for environmental adaptation, which critically introduces dynamic, time-variant relative sensor poses (Fig. \ref{fig:nonrigid_system}). This invalidates classical rigid-body algorithms, presenting significant perception and state estimation challenges \cite{thanabalan2020learning}. In this work, we demonstrate that non-rigid platforms may not necessarily complicate state estimation and sensor fusion. Rather, we show how non-rigid elements and kinetodynamic priors \cite{agudo2015sequential} can even add additional constraints to the system. The leading example we use is monocular motion estimation. While the recovery of scale and inertial alignment are severely ill-posed problems that would normally require fusion with additional sensors \cite{kim2021probabilistic,el2020monocular}, the exploitation of motion and deformation priors perhaps surprisingly renders these dimensions observable.

This work draws inspiration from the work of Li et al. \cite{li2024relative}, which demonstrated the successful embedding of a physical deformation model into a non-rigid multi-perspective camera system. Intriguingly, the deformable connections between sensors in that work acted as a ``passive IMU". Building on this insight, but targeting robust state estimation using a single exteroceptive sensor, this paper addresses a key limitation of prior models. We adopt a mechanical design similar to the Zebedee system \cite{bosse2012zebedee}, where a monocular camera is connected to a mobile platform via a spring mechanism. While this setup serves as a scientific example to explore the potential of priors in non-rigid platforms, we note that there are practically used multi-sensor heads in which non-rigidity was actively added to, for example, enlarge the perceptive field of view \cite{zlot2014three,martinez2019continuous}.

To robustify state estimation under such dynamic non-rigid motion, we explicitly incorporate kinetodynamic priors. This work introduces a novel framework that unifies kinematic and dynamic constraints for non-rigid systems, built upon two core assumptions: 1) Continuous-time Kinematic Models: We use B-Splines to model the smooth motion of at least one platform point, enabling the derivation of high-order derivatives crucial for dynamic analysis. 2) Learned Deformation-force Model: The platform's elastic properties are captured through an injective deformation-force model, efficiently learned via a Multi-Layer Perceptron \cite{li2023mlp}, thus bypassing computationally expensive Finite Element Analysis \cite{lahariya2022learning}. This approach resolves critical challenges like scale and gravity recovery in monocular visual odometry. It continuously applies Newton's Second Law, establishing a physical relationship between the camera's visually derived trajectory acceleration and the acceleration predicted by our learned deformation-force model. Minimizing the discrepancy between these accelerations allows us to determine the unknown scale factor, enabling metric pose and scale estimation from monocular vision \cite{tian2021accurate}. Effectively, any camera motion not explained by the smooth body trajectory is attributed to spring deformations, thus aligning with the support structure's physical properties through an inertial aligning transformation to achieve metric and inertially aligned monocular ego-motion estimation.

Our contributions are summarized as follows:
\begin{itemize}
    \item We introduce compact neural representations for modeling elastic deformation properties of sensor support platforms, coupled with a calibration method using a motion capture device.
    \item We demonstrate how the combination of a suitable body motion model and an elastic deformation model can be leveraged for passive inertial sensing and accurate monocular exteroceptive motion estimation in non-rigid scenarios.
    \item We present the complete computational paradigm, encompassing numerical differentiation of the camera trajectory, variable initialization, and an optimization framework with an embedded, differentiable neural body deformation model.
\end{itemize}

As demonstrated by our results, this outlined setup develops inherent passive inertial sensing capabilities, demonstrating the feasibility of accurate real-world motion estimation from ego-centric video through a mostly mathematical model. Although tested on an exemplary setup, we believe this approach holds significant promise and is applicable to a wide range of future robotic platforms possessing specific motion models and potentially elastic actuation chains.

%% file: sec/main/2_Relatedwork.tex
\section{Related Work}
\label{sec:relatedwork}

\subsection{State Estimation on Non-Rigid Platforms}
The integration of flexible elements into robotic systems, such as deformable UAV wings \cite{hutter2012high} and soft robots \cite{hinzmann2018flexible,hinzmann2019flexible,foehn2022agilicious,hopkins2009survey,liu2005novel}, has gained significant traction. This paradigm contrasts sharply with traditional robotics, which assumes sensors follow rigid trajectories. The dynamic changes in sensor-to-body transformations pose a significant challenge to conventional state estimation algorithms that rely on rigid-body assumptions. A few exceptions exist that explicitly model non-rigid systems, such as the work by Peng et al. \cite{peng2019articulated} and Hinzmann et al. \cite{hinzmann2018flexible,hinzmann2019flexible}. However, these methods are either limited to non-elastic \cite{peng2019articulated} or static scenarios \cite{li2024relative} or depend on overlapping sensor fields of view \cite{hinzmann2019flexible}. Our work, in contrast, tackles the more challenging problem of pure monocular visual state estimation on a non-rigid platform in dynamic environments. The Zebedee system \cite{bosse2012zebedee} is a seminal example that leverages a platform’s elastic motion to extend sensor viewpoints, thereby achieving large-scale 3D coverage, but our approach focuses on using the non-rigidity itself as a source of information.

\subsection{Monocular Visual and Multi-Sensor State Estimation}

Monocular cameras are simple and cheap, but they suffer from a fundamental limitation: scale ambiguity. Classic methods like ORB-SLAM \cite{mur2015orb,mur2017orb,campos2021orb} and COLMAP/GLOMAP \cite{schoenberger2016sfm,schoenberger2016mvs,pan2024glomap,schoenberger2016vote} can reconstruct 3D maps and trajectories, but their scale is always relative. This means the reconstructed environment can be resized by an unknown factor, making it unsuitable for tasks requiring absolute measurements. To overcome this, many systems fuse a camera with other sensors. For example, VINS-Mono \cite{qin2018vins} and maplab \cite{schneider2018maplab} combine a camera with an IMU to resolve scale using inertial data. Similarly, other methods use LiDAR \cite{shan2020lio,shan2018lego,zhang2014loam} or GPS \cite{cao2022gvins}. While effective, these solutions increase hardware cost and complexity. Recently, learning-based methods like Mast3r-SLAM \cite{murai2025mast3r} and VGGt-SLAM \cite{maggio2025vggt} have tried to predict scale directly from images. However, these often require significant computation and cannot be used in a real-time system.

Instead of adding new sensors, we use the non-rigid platform's own elastic properties to resolve scale ambiguity. By modeling the deformation, we achieve metrically accurate state estimation with just a single camera, providing a novel solution for flexible robotics without additional hardware.

\subsection{Passive Sensing and Kinetodynamic Modeling}
Our work is inspired by the concept of ``passive inertial sensing" from Li et al. \cite{li2024relative}, who demonstrated that a sensor's support deformation under gravity can be used to infer inertial information. However, their method was limited to a specific, static physical model. To overcome the limitations of traditional physical models, we employ a more general, data-driven neural model to characterize complex, continuous deformations. While neural networks have been applied to model soft robot control \cite{zheng2020robust} and simulate soft body deformations \cite{wang2024pinn}, these works primarily focus on feed-forward or control problems. In contrast, our approach innovatively integrates a differentiable neural representation of the non-rigid connection's physical properties directly into a monocular visual state estimation framework. This allows us to perform passive perception of continuous deformation states, offering a new perspective for flexible robotics and embodied intelligence research.

%% file: sec/main/3_Methodology.tex
\section{Methodology}
\label{sec:methodology}

\begin{figure*}[t]
  \centering
  \includegraphics[width=2.0\columnwidth]{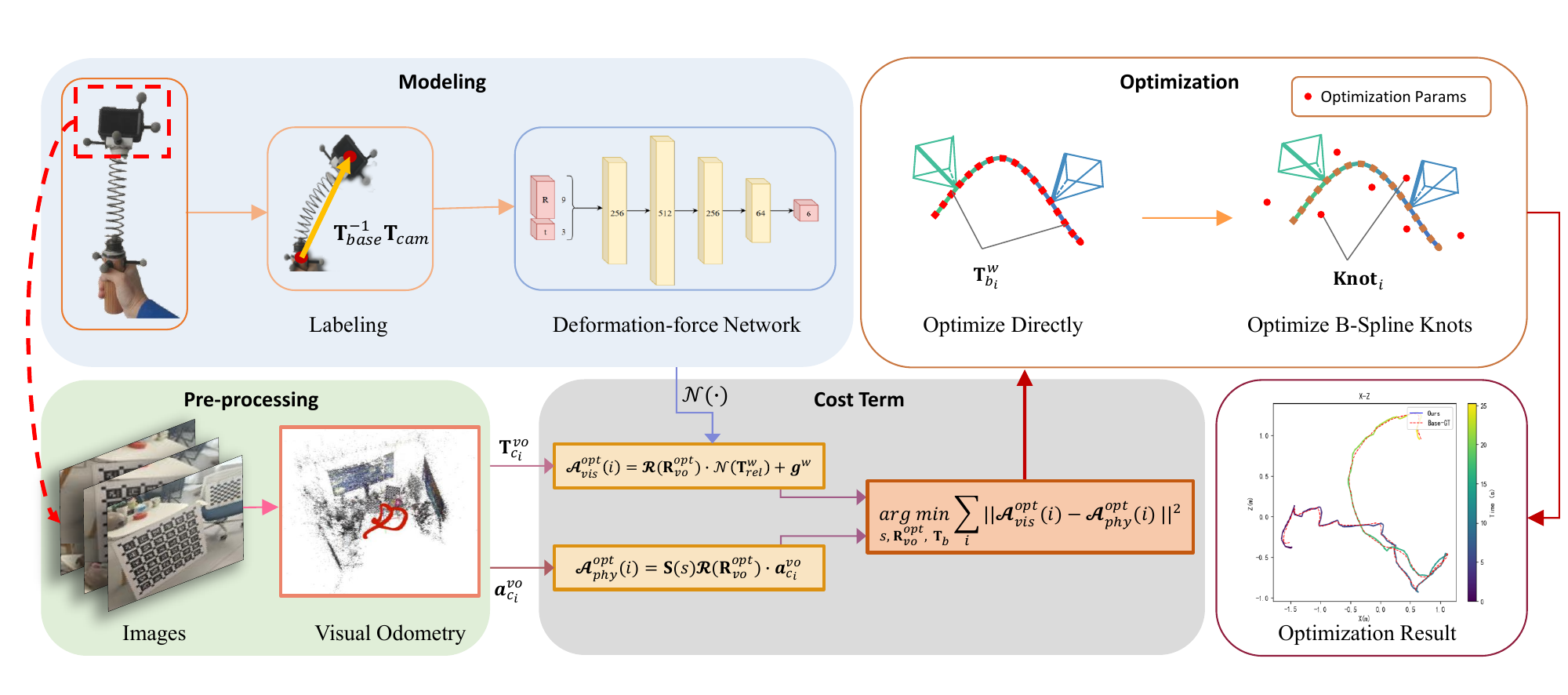}
  \caption{Overview of the proposed pipeline. \textbf{a) Modeling:} A DFN is trained offline to implicitly characterize the non-rigid connection using linear and angular accelerations. \textbf{b) Solving:} We first recover the camera trajectory via visual odometry, followed by a B-Spline-based optimization that minimizes the discrepancy between visual motion and learned physics to estimate the platform's metric state.}
  \label{fig:overview_pipeline}
\end{figure*}

Fig. \ref{fig:overview_pipeline} illustrates our pipeline for motion recovery in a system comprising a camera $c$ and a moving platform $b$ joined by a non-rigid connection. Our approach addresses the coupling of these components in two stages: first, we characterize the elastic link using a Deformation-force Network (DFN); second, we leverage the learned physics prior to recover the metric trajectory and scale of the platform via a B-Spline-based joint optimization.

Throughout this paper, subscripts are used to denote the specific entity (e.g., $c$ for the camera, $b$ for the base), while superscripts indicate the reference coordinate frame in which a quantity is expressed.

\subsection{Preliminaries: B-Spline for $\mathbb{SE}(3)$}

For a $k$-th order B-Spline with control points $\mathbf{T}_i, \mathbf{T}_{i+1}, \dots, \mathbf{T}_N$, the spline is defined as:
\begin{equation}
    \label{bspline}
    \mathbf{T}(t)=\mathbf{T}_i\prod_{j=1}^{k-1}\text{Exp}(\widetilde{\mathbf{B}}_{j}(t)\cdot\text{Log}(\mathbf{T}_{i+j-1}^{-1}\cdot\mathbf{T}_{i+j})),
\end{equation}
where $\text{Exp}(\cdot)$ and $\text{Log}(\cdot)$ represent the exponential and logarithmic mappings on $\mathbb{SE}(3)$, respectively. The basis functions $\widetilde{\mathbf{B}}_{j}(t)$ are determined by the blending matrix $\widetilde{\mathbf{M}}$ with entries:

\begin{equation}
\begin{aligned}
    m_{s,n}^{(k)}=\frac{C_{k-1}^n}{(k-1)!}\sum_{l=s}^{k-1}(-1)^{l-s}C_k^{l-s}(k-1-l)^{k-1-n}\\  s,n\in{0,..,k-1}.
\end{aligned}
\end{equation}

The velocity  can be derived by defining $\mathbf{A}_j(t)=\text{Exp}(\widetilde{\mathbf{B}}_{j}(t))\text{Log}(\mathbf{T}_{i+j-1}^{-1}\cdot\mathbf{T}_{i+j})$, leading to:
\begin{equation}
\label{bspline_acc}
    \dot{\mathbf{T}}(t)=\mathbf{T}_i\cdot\sum_{j=1}^{k-1} \left( \prod_{l=1}^{j-1}\mathbf{A}_l(t) \right) \dot{\mathbf{A}}_j(t) \left( \prod_{l=j+1}^{k-1} \mathbf{A}_l(t) \right).
\end{equation}


Eq. \ref{bspline_acc} yields the linear and angular velocities from the visual odometry trajectory. The corresponding accelerations can be derived in a similar manner by taking the second time derivative, both of which are subsequently used to formulate the optimization objectives.

\subsection{Modeling: Deformation-force Network}
\label{subsec:deformation_force_network}

The mechanical properties of a non-rigid connection can be modeled using a nonlinear differential equation with damping characteristics \cite{friswell2010dynamics}:
\begin{equation}
\boldsymbol{f}_{s}=\kappa_1\Delta \boldsymbol{x} + \kappa_3\Delta \boldsymbol{x}^3 + c\frac{d \Delta \boldsymbol{x}}{d t},
\end{equation}
where $\boldsymbol{f}_{s}$ is the elastic force exerted by the non-rigid connection on the camera. Neglecting air resistance and potential collisions under low-speed conditions, the camera's dynamics are governed by Newton's second law. The relationship between the camera's linear and angular accelerations and its relative displacement is implicitly modeled as:
\begin{equation}
\boldsymbol{a}_{c} - \boldsymbol{g} = H(\Delta \boldsymbol{x}, \Delta\boldsymbol{\theta}),
\end{equation}
where $\boldsymbol{a}_{c} = \begin{bmatrix} {\boldsymbol{a}_{c}^{linear}}^T & {\boldsymbol{a}_{c}^{angular}}^T \end{bmatrix}^T$ is the 6-DoF acceleration and $\boldsymbol{g} = [0, 0, -g, 0, 0, 0]^T$ represents gravity. The relative pose $\mathbf{T}_{b}^{-1}\mathbf{T}_{c}$ captures the structural deformation. Thus, we approximate this mapping using a neural network $\mathcal{N}$:
\begin{equation}
    \boldsymbol{a}_{c} - \boldsymbol{g} = \mathcal{N}(\mathbf{T}_{b}^{-1}\mathbf{T}_{c}).
\end{equation}

To ensure the network learns physics in a consistent sensor-centric frame, we supervise it using ground-truth motion projected into the camera frame $c$:
\begin{equation}
\label{eq:normlization}
\boldsymbol{a}^{c}_c - \boldsymbol{g}^{c} = \mathbfcal{R}(\mathbf{R}_{c}^{gt_i})^{-1}(\boldsymbol{a}^{gt_i}_c - \boldsymbol{g}^{gt_i}_c),
\end{equation}
where $\mathbfcal{R}(\mathbf{R}) = \text{diag}(\mathbf{R}, \mathbf{R})$ is a $6\times6$ block-diagonal rotation matrix, and ${gt_i}$ denotes the ground-truth inertial frame. The ground-truth kinematic data is captured via an external motion capture system and is exclusively utilized during the offline training phase.

\subsection{Optimization: State Estimation and Scale Recovery}
\label{sec:Scale recovery and state estimation}

Visual odometry (VO) provides unscaled camera trajectories. We relate these to the optimized metric trajectories via a rotation $\mathbf{R}_{vo}^{opt}$ and a scaling factor $s$. Since angular motion is scale-invariant, we define the scaling matrix $\mathbf{S}(s) = \text{diag}(s, s, s, 1, 1, 1)$. The visual acceleration constraint is thus:
\begin{equation}
\label{eq:acc-}
\boldsymbol{a}_{c_i}^{opt} = \mathbf{S}(s) \mathbfcal{R}(\mathbf{R}_{vo}^{opt}) \boldsymbol{a}_{c_i}^{vo}.
\end{equation}

Physically, the optimized acceleration $\boldsymbol{a}_{c_i}^{opt}$ must also satisfy the dynamics predicted by the network. By transforming the network output to the $opt$ frame, we obtain:

\begin{equation}
\boldsymbol{a}_{c_i}^{opt} =\mathbfcal{R}(\mathbf{R}_{c_i}^{opt}) \mathcal{N} ((\mathbf{T}^{opt}_{b_i})^{-1} \mathbf{T}_{c_i}^{opt}) + \boldsymbol{g}^{opt}.
\end{equation}

Defining the residual $\boldsymbol{r}_i$ as the discrepancy between physical prediction and visual observation:
\begin{equation}
\boldsymbol{r}_i = \mathbfcal{R}(\mathbf{R}_{c_i}^{opt}) \mathcal{N} ((\mathbf{T}^{opt}_{b_i})^{-1} \mathbf{T}_{c_i}^{opt}) + \boldsymbol{g}^{opt} - \mathbf{S}(s) \mathbfcal{R}(\mathbf{R}_{vo}^{opt}) \boldsymbol{a}_{c_i}^{vo}.
\end{equation}

Here, we directly set $\boldsymbol{g}^{opt} = \boldsymbol{g}^{w}$ and then verify $\mathbf{R}_{vo}^{opt}$ to ensure the angle between the optimized coordinate system and the ground truth gravity direction remains small.

To ensure efficiency and smoothness, we parameterize the platform trajectory using B-Spline control points $\mathbf{T}^{opt}_i$. The joint optimization objective is:
\begin{equation}
\label{opt-bspline}
\mathop{\arg\min}_{s,\mathbf{T}_{0:n}^{opt},\mathbf{R}_{vo}^{opt}} \sum_i \left\| \boldsymbol{r}_i \right\|^2.
\end{equation}
This formulation enables the recovery of the metric scale $s$ by aligning the dimensionless visual motion with the metric forces learned from structural deformations.

%% file: sec/main/4_Experiments.tex
\section{Experiments}
\label{sec:experiments}

\subsection{Implementation Details}
\label{sec:implementation}

\begin{figure}[htpb]
  \centering
  \vspace{-0.3 cm}
  \includegraphics[width=1.0\columnwidth]{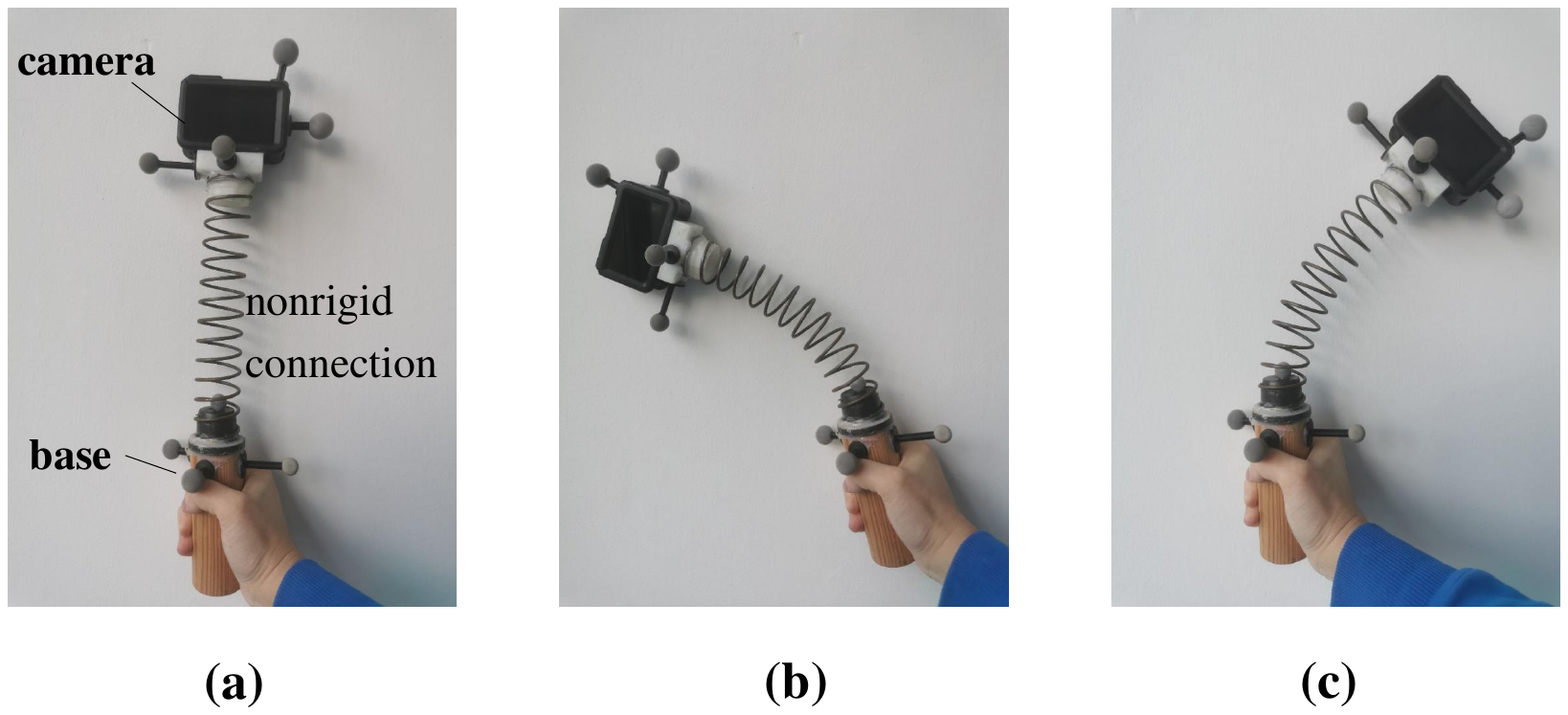}
  \vspace{-2.3 cm}
  \caption{Hardware setup. The system comprises a base, a camera, and a non-rigid spring connection. The optimization targets are the pose and scale of the base.}
  \label{fig:real_exp_setup}
\end{figure}

\subsubsection{Hardware and Data Acquisition}
As illustrated in Fig. \ref{fig:real_exp_setup}, the experimental non-rigid system consists of a monocular camera attached to a moving base via a passive elastic spring. To capture irregular motion induced by body-induced accelerations and vibrations, a handheld holder is utilized. The camera's 6-DoF motion is governed by the interaction between the spring's restoration force and gravity.

Ground truth trajectories are acquired using an optical motion capture system. Markers are attached to both the camera and the holder (base). The coordinate systems are gravity-aligned to ensure consistency with the gravity-aware network training and optimization. Trajectory sequences, ranging from 25 to 45 seconds, were recorded for evaluation.

\subsubsection{Optimization Setup}
We employ COLMAP \cite{schoenberger2016sfm} for robust visual odometry. As described in Eq. \ref{opt-bspline}, the optimization terms for the base trajectory and scale are constructed using the camera's trajectory and acceleration derived from VO. The problem is solved using the Ceres Solver.

\subsubsection{Evaluation Metrics}
\begin{figure*}[htpb]
  \centering
  \includegraphics[width=2.0\columnwidth]{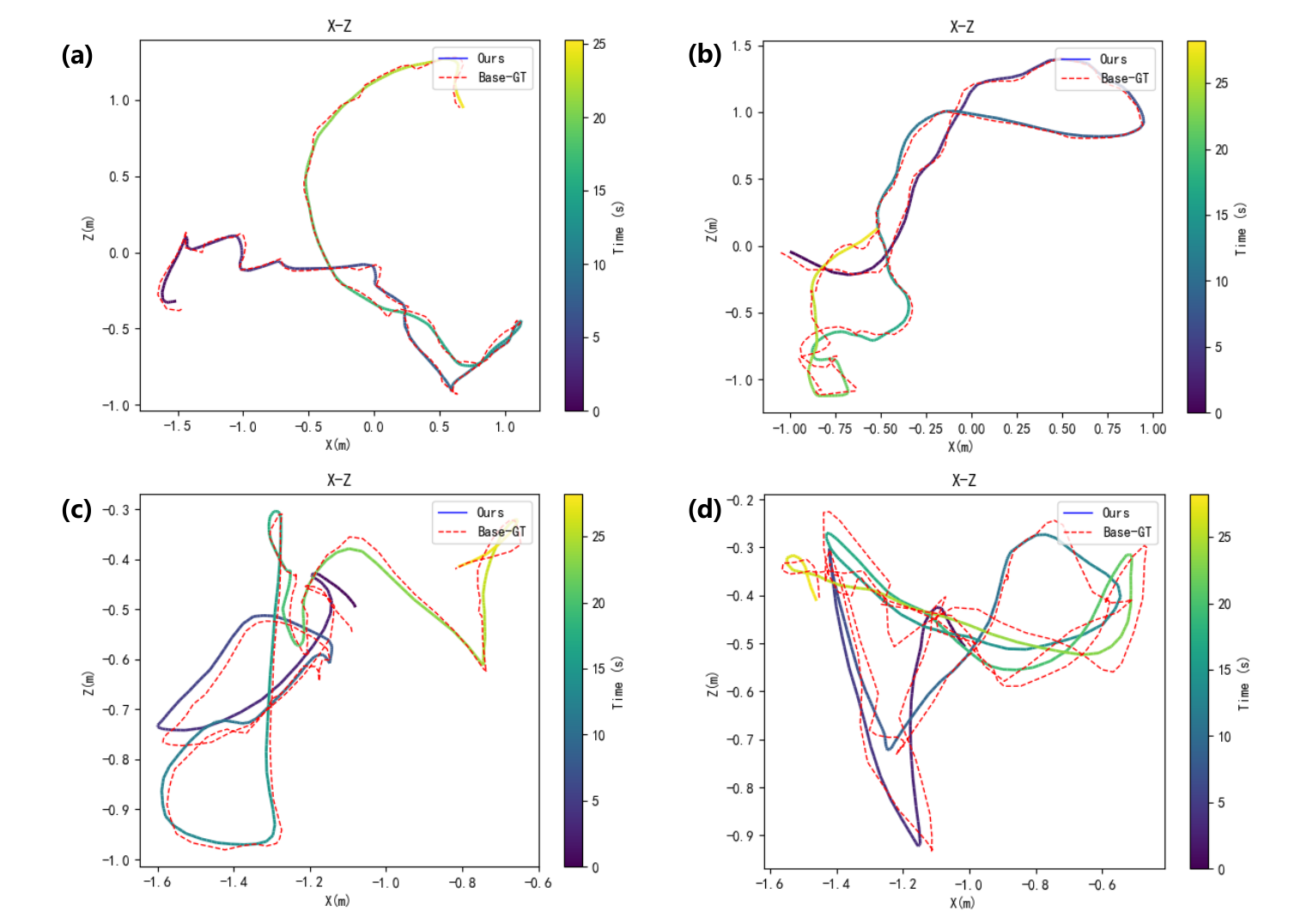}
  \caption{Qualitative results. 2D X-Z projections of ground truth (solid) and optimized (dashed) trajectories are shown. The alignment demonstrates the effectiveness of the proposed method across various motion patterns. (d) illustrates a challenging case with motion blur.}
  \label{fig:optimization_results}
\end{figure*}

\label{sec:metrics}
Performance is evaluated using three key metrics:
\begin{itemize}
    \item \textbf{Trajectory Accuracy:} We compute the Absolute Pose Error (APE) for both the VO output ($\mathbf{T}_{c_i}^{vo}$ vs. $\mathbf{T}_{c_i}^{gt}$) and the optimized base trajectory ($\mathbf{T}_{b_i}^{w}$ vs. $\mathbf{T}_{b_i}^{gt}$).
    \item \textbf{Scale Accuracy:} The relative scale error is defined as $err_s = \frac{\|s_{gt} - s\|}{s_{gt}}$, where $s_{gt}$ is the ground truth scale aligned using EVO \cite{grupp2017evo}.
    \item \textbf{Gravity Alignment:} Since physical gravity is integrated into our pipeline, we evaluate the angular error between the optimized gravity direction and the physical gravity vector: $err_G=\arccos(\frac{\boldsymbol{g}^{opt}\cdot \boldsymbol{g}^{gt}}{\|\boldsymbol{g}^{opt}\|\cdot \|\boldsymbol{g}^{gt}\|})$.
\end{itemize}

\subsection{Experimental Results}

\subsubsection{Simulation Experiments}
To assess robustness against noise and outliers, we generated synthetic camera trajectories by applying Gaussian noise and outliers to real-world ground truth data. The simulated pose $\mathbf{T}_{c_i}^{sim}$ is defined as:

\begin{equation}
    \mathbf{T}_{c_i}^{sim} = \mathbf{T}_{c_i}^{gt} \oplus (\sigma \cdot \boldsymbol{\xi}) ,  \quad \boldsymbol{\xi} \sim \mathcal{N}(\mathbf{0}, \mathbf{I}),
\end{equation}

where $\sigma$ represents the noise amplitude. Additionally, to simulate gross errors, a proportion of the trajectory corresponding to the outlier ratio $\mathbf{O}_r$ is replaced with random transformations. We conducted experiments across varying noise levels (0\% to 10\%) and outlier ratios (0\% to 5\%). For each setting, results were averaged over six independent runs.

\begin{table}[hbtp]
  \centering
  \caption{Average performance metrics under varying noise magnitudes.}
    \begin{tabular}{cccccc}
    \toprule
    \multirow{2}[4]{*}{Noise} & \multicolumn{3}{c}{APE} & \multirow{2}[4]{*}{$\text{err}_s$} & \multirow{2}[4]{*}{$\text{err}_G$ (\degree)} \\
\cmidrule{2-4}          & Mean (m) & Median (m) & STD   &       &  \\
    \midrule
    0\%   & 0.036 & 0.035 & 0.014 & 0.006 & 0.447 \\
    3\%   & 0.040 & 0.038 & 0.043 & 0.003 & 0.638 \\
    5\%   & 0.067 & 0.062 & 0.034 & 0.058 & 1.328 \\
    10\%  & 0.096 & 0.092 & 0.041 & 0.042 & 1.841 \\
    \bottomrule
    \end{tabular}%
  \label{tab:simulation_noise}
\end{table}%

\begin{table}[hbtp]
  \centering
  \caption{Average performance metrics under varying outlier proportions (fixed 3\% noise).}
    \begin{tabular}{cccccc}
    \toprule
    \multirow{2}[4]{*}{Outlier} & \multicolumn{3}{c}{APE} & \multirow{2}[4]{*}{$\text{err}_s$} & \multirow{2}[4]{*}{$\text{err}_G$ (\degree)} \\
\cmidrule{2-4}          & Mean (m) & Median (m) & STD   &       &  \\
    \midrule
    0\%   & 0.040 & 0.038 & 0.043 & 0.003 & 0.638  \\
    1\%   & 0.083 & 0.080 & 0.039 & 0.062 & 1.579 \\
    3\%   & 0.141 & 0.138 & 0.042 & 0.065 & 1.943 \\
    5\%   & 0.164 & 0.158 & 0.061 & 0.108 & 2.657 \\
    \bottomrule
    \end{tabular}%
  \label{tab:simulation_outlier}%
\end{table}%

Table \ref{tab:simulation_noise} demonstrates the method's robustness to amplitude noise. Even at a 10\% noise level, scale and gravity errors remain low. Furthermore, as shown in Table \ref{tab:simulation_outlier}, the algorithm maintains acceptable accuracy even when challenged with up to 5\% outliers, validating its stability in non-ideal conditions.

\subsubsection{Real-World Experiments}

In real-world scenarios, we utilize the estimated $\mathbf{T}_{c_i}^{vo}$ and acceleration $\mathbf{a}_{c_i}^{vo}$ from COLMAP to optimize the base trajectory. 
In our framework, Visual Odometry is treated as an input component, providing initial kinematic constraints for the subsequent optimization. To the best of our knowledge, no other established methods effectively address trajectory estimation specifically for such non-rigid systems.

Fig. \ref{fig:optimization_results} visualizes the optimization results. The strong overlap between estimated and ground truth trajectories confirms the accuracy of our pipeline. 
However, large system deformations can induce rapid camera motion, causing motion blur that degrades VO performance. This is observed in Fig. \ref{fig:optimization_results}(d), where accuracy is slightly reduced compared to other sequences. This limitation could be mitigated by using high-frame-rate cameras or ensuring a feature-rich environment.

Quantitative results are detailed in Table \ref{tab:error_analysis_1}. The analysis confirms that our method successfully recovers the metric scale and estimates the base trajectory of a non-rigid system using only a monocular camera, without requiring additional sensors.

\subsection{Discussion}
\label{sec:discussion}

The performance discrepancy between simulation and real-world experiments primarily stems from the quality of the input camera trajectory. While simulations utilize ground truth with additive Gaussian noise to isolate algorithmic efficacy, real-world experiments rely on visual odometry for state estimation. Crucially, the non-rigid spring coupling induces high-frequency vibrations and rapid camera motion, resulting in significant motion blur. This degradation in image quality inevitably impairs the VO tracking accuracy. Since our proposed pipeline depends on the estimated camera kinematics, errors from the VO stage propagate to the final base trajectory optimization, thereby limiting the overall system precision in dynamic real-world scenarios compared to the idealized simulation environment.

%% file: sec/main/6_Ablation.tex
\section{Ablation Study}

\begin{table*}[htbp]
  \centering
  \vspace{0.3 cm}
  \caption{Quantitative error analysis on real-world experimental data. APE (Optimization) measures the error between optimized base pose $\mathbf{T}_{b_i}^w$ and ground truth $\mathbf{T}_{b_i}^{gt}$; APE (VO Component) measures the error between visual odometry $\mathbf{T}_{c_i}^{vo}$ and ground truth $\mathbf{T}_{c_i}^{gt}$. Note that the VO trajectory is Sim(3)-aligned to the ground truth. Utilizing this oracle scale may occasionally make VO translation errors lower than our optimized results.}
    \begin{tabular}{cccccccccccc}
    \toprule
    \multirow{2}[4]{*}{Seq.} & \multicolumn{3}{c}{APE (Optimization)} &       & \multicolumn{3}{c}{APE (VO Component)} & \multirow{2}[4]{*}{$\text{err}_s$} & \multirow{2}[4]{*}{$s$} & \multirow{2}[4]{*}{$s_{gt}$} & \multirow{2}[4]{*}{Gravity Err. (\degree)}\\
\cmidrule{2-4}\cmidrule{6-8}          & Mean(m) & Median(m) & STD   &       & Mean (m) & Median (m) & STD   &       &       &       &  \\
    \midrule
    1     & 0.120 & 0.081 & 0.088 &       & 0.196 & 0.201 & 0.086 & 0.271 & 0.183 & 0.251 & 7.34 \\
    2     & 0.149 & 0.138 & 0.069 &       & 0.158 & 0.147 & 0.086 & 0.246 & 0.308 & 0.247 & 6.67 \\
    3     & 0.170 & 0.121 & 0.039 &       & 0.159 & 0.154 & 0.086 & 0.201 & 0.194 & 0.243 & 4.83 \\
    4     & 0.136 & 0.101 & 0.068 &       & 0.274 & 0.273 & 0.156 & 0.147 & 0.132 & 0.115 & 3.28 \\
    5     & 0.181 & 0.188 & 0.093 &       & 0.127 & 0.117 & 0.077 & 0.276 & 0.143 & 0.112 & 6.28 \\
    6     & 0.165 & 0.172 & 0.064 &       & 0.161 & 0.132 & 0.117 & 0.707 & 0.031 & 0.106 & 13.4 \\
    7     & 0.158 & 0.152 & 0.081 &       & 0.234 & 0.212 & 0.112 & 0.669 & 0.202 & 0.121 & 6.88 \\
    8     & 0.150 & 0.156 & 0.081 &       & 0.306 & 0.251 & 0.211 & 0.170 & 0.112 & 0.135 & 7.10 \\
    9     & 0.283 & 0.257 & 0.148 &       & 0.289 & 0.255 & 0.165 & 0.622 & 0.318 & 0.196 & 6.82 \\
    10    & 0.213 & 0.154 & 0.143 &       & 0.268 & 0.238 & 0.141 & 0.711 & 0.344 & 0.201 & 8.22 \\
    11    & 0.218 & 0.211 & 0.094 &       & 0.292 & 0.293 & 0.155 & 0.372 & 0.177 & 0.129 & 7.12 \\
    12    & 0.293 & 0.254 & 0.142 &       & 0.248 & 0.240 & 0.128 & 0.476 & 0.056 & 0.107 & 2.12 \\
    13    & 0.715 & 0.598 & 0.406 &       & 0.563 & 0.584 & 0.217 & 1.509 & 0.527 & 0.210  & 5.98 \\
    14    & 0.120 & 0.081 & 0.088 &       & 0.121 & 0.110 & 0.072 & 0.648 & 0.183 & 0.521 & 7.34 \\
    15    & 0.461 & 0.399 & 0.307 &       & 0.285 & 0.294 & 0.171 & 0.027 & 0.258 & 0.251 & 0.92 \\
    16    & 0.094 & 0.084 & 0.042 &       & 0.096 & 0.095 & 0.045 & 0.671 & 0.028 & 0.085 & 7.27 \\
    \midrule
    Median & 0.167 & 0.155 & - &       & 0.241 & 0.225 & - & 0.483 & -     & -     & 6.85 \\
    Mean  & 0.226 & 0.196 & - &       & 0.236 & 0.224 & - & 0.424 & -     & -     & 6.36 \\
    \bottomrule
    \end{tabular}
  \label{tab:error_analysis_1}
\end{table*}

\begin{table}[htbp]
  \centering
  \caption{Comparison of errors: Normalized (with Eq. \ref{eq:normlization}) vs. Directly (without Eq. \ref{eq:normlization}). We specifically report metrics such as the mean and median errors for linear and angular accelerations across each axis.}
    \begin{tabular}{rrrrr}
    \toprule
          &       & \multicolumn{1}{c}{Mean} & \multicolumn{1}{c}{Median} & \multicolumn{1}{c}{STD} \\
    \midrule
    \multicolumn{1}{c|}{\multirow{6}[0]{*}{$E_{normalized}$}} & \multicolumn{1}{c}{$^{(x)}{a}_{c}^{linear}$ ($m/s^2$)} & \multicolumn{1}{c}{0.323} & \multicolumn{1}{c}{0.327} & \multicolumn{1}{c}{0.151} \\
    \multicolumn{1}{c|}{} & \multicolumn{1}{c}{$^{(y)}a_{c}^{linear}$ ($m/s^2$)} & \multicolumn{1}{c}{0.799} & \multicolumn{1}{c}{0.653} & \multicolumn{1}{c}{0.636} \\
    \multicolumn{1}{c|}{} & \multicolumn{1}{c}{$^{(z)}a_{c}^{linear}$ ($m/s^2$)} & \multicolumn{1}{c}{0.167} & \multicolumn{1}{c}{0.147} & \multicolumn{1}{c}{0.109} \\
    \multicolumn{1}{c|}{} & \multicolumn{1}{c}{$^{(x)}a_{c}^{angular}$ ($rad/s^2$)} & \multicolumn{1}{c}{0.148} & \multicolumn{1}{c}{0.119} & \multicolumn{1}{c}{0.122} \\
    \multicolumn{1}{c|}{} & \multicolumn{1}{c}{$^{(y)}a_{c}^{angular}$ ($rad/s^2$)} & \multicolumn{1}{c}{0.121} & \multicolumn{1}{c}{0.092} & \multicolumn{1}{c}{0.109} \\
    \multicolumn{1}{c|}{} & \multicolumn{1}{c}{$^{(z)}a_{c}^{angular}$ ($rad/s^2$)} & \multicolumn{1}{c}{0.203} & \multicolumn{1}{c}{0.168} & \multicolumn{1}{c}{0.171} \\
    \midrule
    \multicolumn{1}{c|}{\multirow{6}[0]{*}{$E_{directly}$}} & \multicolumn{1}{c}{$^{(x)}a_{c}^{linear}$ ($m/s^2$)} & \multicolumn{1}{c}{1.128} & \multicolumn{1}{c}{0.91} & \multicolumn{1}{c}{0.898} \\
    \multicolumn{1}{c|}{} & \multicolumn{1}{c}{$^{(y)}a_{c}^{linear}$ ($m/s^2$)} & \multicolumn{1}{c}{1.578} & \multicolumn{1}{c}{1.197} & \multicolumn{1}{c}{1.213} \\
    \multicolumn{1}{c|}{} & \multicolumn{1}{c}{$^{(z)}a_{c}^{linear}$ ($m/s^2$)} & \multicolumn{1}{c}{1.251} & \multicolumn{1}{c}{1.028} & \multicolumn{1}{c}{1.089} \\
    \multicolumn{1}{c|}{} & \multicolumn{1}{c}{$^{(x)}a_{c}^{angular}$ ($rad/s^2$)} & \multicolumn{1}{c}{0.456} & \multicolumn{1}{c}{0.401} & \multicolumn{1}{c}{0.338} \\
    \multicolumn{1}{c|}{} & \multicolumn{1}{c}{$^{(y)}a_{c}^{angular}$ ($rad/s^2$)} & \multicolumn{1}{c}{0.353} & \multicolumn{1}{c}{0.296} & \multicolumn{1}{c}{0.283} \\
    \multicolumn{1}{c|}{} & \multicolumn{1}{c}{$^{(z)}a_{c}^{angular}$ ($rad/s^2$)} & \multicolumn{1}{c}{0.405} & \multicolumn{1}{c}{0.377} & \multicolumn{1}{c}{0.261} \\
    \bottomrule
    \end{tabular}%
  \label{tab:unnormalized-cmp}%
\end{table}%

To validate the importance of our data collection and processing pipeline for accurately modeling a non-rigid system, we design two ablation studies. In the first study, we eliminate the step of normalizing data to the camera coordinate system. The second study demonstrates the advantage of using multi-dimensional motion patterns for robustly modeling gravity.

\subsubsection{Relative Pose Normalization (Eq. \ref{eq:normlization})}
To demonstrate the importance of normalizing data to the camera coordinate system, we reconstructed the dataset and performed system identification again, this time removing the step represented by Eq. \ref{eq:normlization}. Table \ref{tab:unnormalized-cmp} shows the resulting modeling errors, listing the acceleration and angular acceleration errors in each axis. These results highlight the significant advantage of our normalization approach.

\subsubsection{Movement Patterns on Modeling}
As previously mentioned, we deliberately curated a diverse set of motion patterns for our training data. This strategy was employed to ensure the final model could accurately handle the constant-direction gravitational vector. Table \ref{tab:patt-alb} presents the results of training with these distinct patterns. The results indicate that while individual patterns enable the network to learn dynamics specific to certain axes, a combination of all patterns is crucial for the network to generalize and robustly handle all motion scenarios.

%% file: sec/main/5_Conclusion.tex
\section{Conclusion and Discussions}
\label{sec:conclusion}

\begin{table}[t]
  \centering
  \caption{Modeling performance with different motion patterns and the units are $m/s^2$ and $rad/s^2$, respectively. The best and second-best performing metrics are bolded for clarity. (Pattern A: Pure translation, B: Pure rotation, C: Translation + vertical, D: Rotation + vertical, detailed in Appendix A.)}
    \begin{tabular}{lrrrrr}
    \toprule
          & \multicolumn{1}{l}{Pattern A} & \multicolumn{1}{l}{Pattern B} & \multicolumn{1}{l}{Pattern C} & \multicolumn{1}{l}{Pattern D} & \multicolumn{1}{l}{Total} \\
    \midrule
    $^{(x)}a_{c}^{linear}$  & 1.071 & \textbf{0.864} & 1.198 & 1.041 & \textbf{0.939} \\
    $^{(y)}a_{c}^{linear}$  & 5.244 & 4.816 & 4.808 & \textbf{4.186} & \textbf{2.325} \\
    $^{(z)}a_{c}^{linear}$  & \textbf{0.534} & 1.781 & 0.665 & 0.834 & \textbf{0.485} \\
    $^{(x)}a_{c}^{angular}$  & \textbf{0.368} & 0.375 & \textbf{0.348} & 0.400   & 0.402 \\
    $^{(y)}a_{c}^{angular}$  & 0.376 & \textbf{0.355} & 0.455 & 0.436 & \textbf{0.329} \\
    $^{(z)}a_{c}^{angular}$  & \textbf{0.346} & \textbf{0.353} & 0.449 & 0.447 & 0.552 \\
    \bottomrule
    \end{tabular}%
  \label{tab:patt-alb}%
\end{table}%

This paper introduced a novel perception system leveraging a monocular camera passively coupled to a moving platform via a non-rigid connection. By integrating a learned deformation-force model with our state estimation framework, we demonstrated that non-rigid coupling provides crucial constraints for passive inertial sensing, effectively resolving inherent scale and inertial alignment ambiguities. Our experiments validated that correctly modeling kineto-dynamics enables accurate motion and metric scale estimation.

A fundamental limitation of our system is its reliance on the coupling’s elasticity; as the connection approaches ideal rigidity, the informative deformation cues diminish, and the system's advantage degenerates to standard monocular setups. Additionally, the current batch optimization framework entails increasing computational overhead over long trajectories. We also observed that rotational precision, while competitive, remains more sensitive to optimization on the $\mathbb{SO}(3)$ manifold compared to translation. Future work will focus on implementing a sliding-window optimization for real-time efficiency and investigating manifold-aware loss functions to further refine rotational accuracy. This approach marks a significant step toward robust state estimation for future robotic platforms with elastic actuation chains.

%% file: sec/main/7_Acknowledgement.tex
\section{Acknowledgement}

The authors would like to thank the fund support from Natural Science Foundation of China(W2531052) and the 3D printing support from ShanghaiTech SIST Machine Shop on real experiment setup.

%% file: sec/appendix/appendix_icra.tex
\appendix

\subsection{Data Acquisition}
\setcounter{figure}{0} 
\renewcommand{\thefigure}{A\arabic{figure}} 
\setcounter{table}{0} 
\renewcommand{\thetable}{A\arabic{table}} 
\setcounter{equation}{0}
\renewcommand{\theequation}{A\arabic{equation}} 
\label{appendix-da}

\begin{figure}[htbp]
    \centering
    \includegraphics[width=1\linewidth]{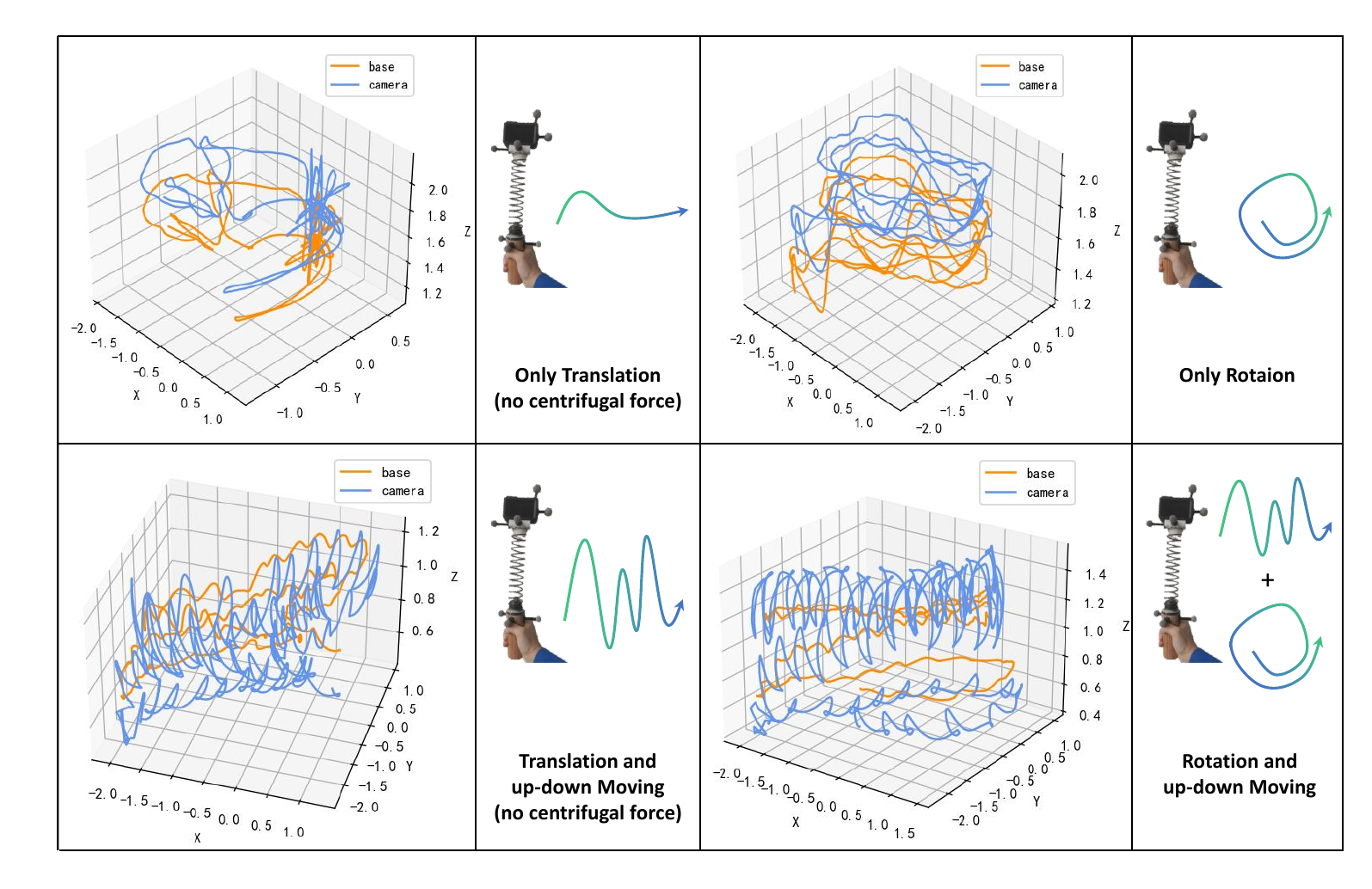}
    \caption{Trajectories for different states of motion.}
    \label{fig:data-fig1}
\end{figure}

We use a tracking system to acquire ground truth trajectories of both the camera and the base at 360 Hz. We collected trajectories lasting 25-45 seconds across various patterns, encompassing both smooth and vigorous motions with diverse poses. As illustrated in Fig. \ref{fig:data-fig1}, for each pose, we collected data for:

\begin{itemize}
    \item \textbf{Pure translational motion}: The spring primarily undergoes tangential deformation along the direction of motion.
    \item \textbf{Pure rotational motion}: The camera experiences radial stretching of the spring due to centrifugal force.
    \item \textbf{Translational motion with vertical movement (gravity aligned)}: This setup amplifies the effect of gravity, enabling accurate modeling of its influence.
    \item \textbf{Rotational motion with vertical movement (gravity aligned)}: This case induces the most complex spring deformation, resulting from the combined effects of centrifugal force and gravity.
\end{itemize}

\begin{figure}[htbp]
    \centering
    \includegraphics[width=1\linewidth]{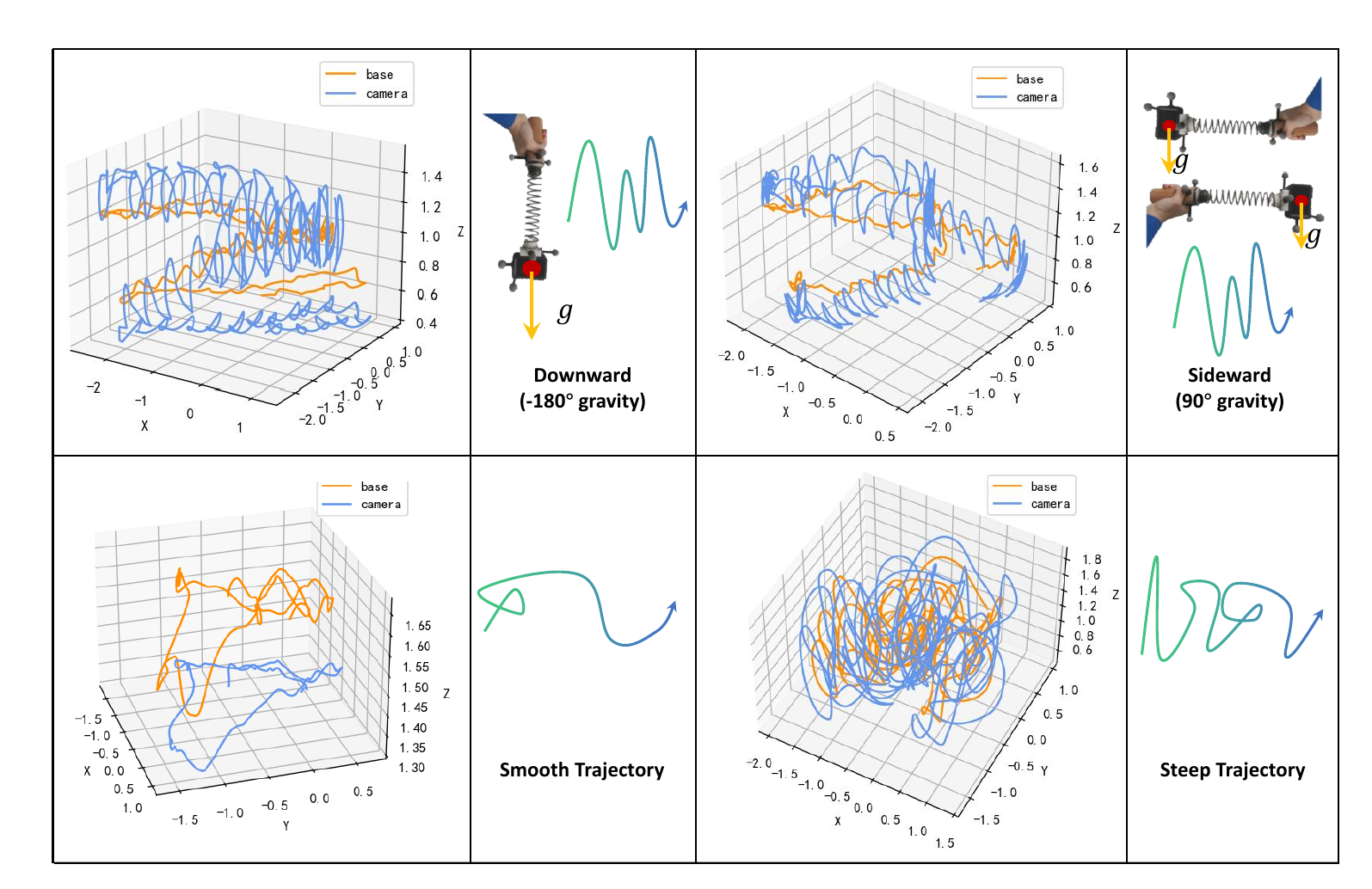}
    \caption{Trajectories with varying gravitational orientations and smoothness levels.}
    \vspace{-0.7 cm}
    \label{fig:data-fig2}
\end{figure}

To capture the spring's behavior under diverse gravitational influences, we collected data with the system inverted and on its side, addressing the compressive effect of gravity in a vertical downward orientation. Fig. \ref{fig:data-fig2} illustrates these varied configurations, alongside trajectories demonstrating different intensities of complex motion states.

\subsection{Training DFN}
\setcounter{figure}{0} 
\renewcommand{\thefigure}{B\arabic{figure}} 
\setcounter{table}{0} 
\renewcommand{\thetable}{B\arabic{table}} 
\setcounter{equation}{0}
\renewcommand{\theequation}{B\arabic{equation}} 

We employ an MLP to model the non-rigid connection. The network's computational complexity and parameter count were quantified using the \texttt{thop} library, revealing 375.392 KFLOPs and 377.222 K parameters, respectively.

We use approximately 259 K samples derived from 16 collected trajectories for training, and another 16 sequences reserved for real experiments. These samples are partitioned into training, testing, and validation sets in a 7:2:1 ratio. We use Adam as an optimizer, and the learning rate is 1e-4. We train the network for 100 epochs with the batch size 1024.

\subsection{Observability Analysis of the Metric Scale}
\setcounter{figure}{0} 
\renewcommand{\thefigure}{C\arabic{figure}} 
\setcounter{table}{0} 
\renewcommand{\thetable}{C\arabic{table}} 
\setcounter{equation}{0}
\renewcommand{\theequation}{C\arabic{equation}} 

The metric scale $s$ is inherently unobservable in pure monocular visual odometry due to the scale-invariant nature of perspective projection. In our framework, this scale ambiguity is resolved through the macroscopic coupling of visual kinematics and learned physical dynamics.

Let $\mathbf{a}_{ci}^{vo,linear}$ be the dimensionless linear acceleration derived from visual odometry. The physical consistency residual for the linear acceleration component can be expressed as a function of the scale $s$:
\begin{equation}
    \mathbf{r}_i^{linear}(s) = \mathbf{R}_{ci}^{opt}\mathcal{N}_{linear}\left( \mathbf{T}_{c_i}^{b_i}(s) \right) + \mathbf{g}^{w} - s \mathbf{R}_{vo}^{opt}\mathbf{a}_{ci}^{vo,linear},
    \tag{C1}
    \label{eq:scale_residual}
\end{equation}
where $\mathcal{N}_{linear}(\cdot)$ extracts the linear acceleration prediction from the Deformation-force Network, and $\mathbf{g}^w$ represents the physical gravity vector. 

The network $\mathcal{N}$ is trained offline using ground-truth kinematics captured by a motion capture system. This methodology ensures that $\mathcal{N}$ intrinsically operates in a true metric space. It acts as a learned kinetodynamic prior, mapping relative structural deformations directly to absolute metric forces and accelerations, thereby embedding the physical elasticity scale into the model's weights.

And the formulation explicitly incorporates the physical gravity vector $\mathbf{g}^{w}$ as an absolute metric reference, which inherently prevents any potential linear scaling equivalence between the kinematics and the dynamics. Specifically, if an erroneous, arbitrary scaling factor $\alpha \neq 1$ were applied to the spatial dimensions of the system, the purely kinematic visual acceleration would scale proportionally. However, the physical acceleration prediction would not scale equivalently due to the constant additive gravity term:
\begin{equation}
    \alpha s \mathbf{R}_{vo}^{opt} \mathbf{a}_{ci}^{vo, linear} \neq \mathbf{R}_{ci}^{opt}\mathcal{N}_{linear}\left( \mathbf{T}_{c_i}^{b_i}(\alpha s) \right) + \mathbf{g}^w.
    \tag{C2}
\end{equation}
This inequality demonstrates that an arbitrary scale cannot arbitrarily cancel out across the dynamic constraints. As a result, the optimal scale $s$ is uniquely determined and aligned to the absolute metric scale defined by the gravitational constant.